\documentclass[runningheads]{llncs}
\usepackage{graphicx}
\usepackage{amsmath,amssymb} 
\usepackage{color}
\usepackage{xspace}
\usepackage{gensymb}
\usepackage{multirow}
\usepackage{makecell}
\usepackage[caption=false]{subfig}
\usepackage[width=122mm,left=12mm,paperwidth=146mm,height=193mm,top=12mm,paperheight=217mm]{geometry}
\usepackage[activate={true,nocompatibility},final,tracking=true,kerning=true,spacing=true,factor=1100,stretch=10,shrink=10]{microtype}
\begin{document}
\pagestyle{headings}
\mainmatter

\title{Deep Autoencoder for Combined Human Pose Estimation and Body Model Upscaling} 

\titlerunning{Deep Autoencoder for Human Pose Estimation and Body Upscaling}

\authorrunning{M. Trumble, A Gilbert, A Hilton \& J Collomosse}

\author{Matthew Trumble$^{1}$, Andrew Gilbert$^{1}$, Adrian Hilton$^{1}$, John Collomosse$^{1,2}$}


\institute{$^{1}$Centre for Vision Speech and Signal Processing,\\
	University of Surrey\\
	$^{2}$Creative Intelligence Lab, \\
	Adobe Research\\
}

\newcommand{\etal}{{\em et~al.}\xspace}
\newcommand{\eg}{e.\,g.\xspace}
\newcommand{\ie}{i.\,e.\xspace}
\newcommand{\squeezeup}{\vspace{-4mm}}

\maketitle

\begin{abstract}
We present a method for simultaneously estimating 3D human pose and body shape from a sparse set of wide-baseline camera views.  We train a symmetric convolutional autoencoder with a dual loss that enforces learning of a latent representation that encodes skeletal joint positions, and at the same time learns a deep representation of volumetric body shape.  We harness the latter to up-scale input volumetric data by a factor of $4 \times$, whilst recovering a 3D estimate of joint positions with equal or greater accuracy than the state of the art.  Inference runs in real-time (25 fps) and has the potential for passive human behavior monitoring where there is a requirement for high fidelity estimation of human body shape and pose.

\end{abstract}
\squeezeup
\squeezeup
\squeezeup
\begin{figure}
\centering
\includegraphics[width=0.8\linewidth]{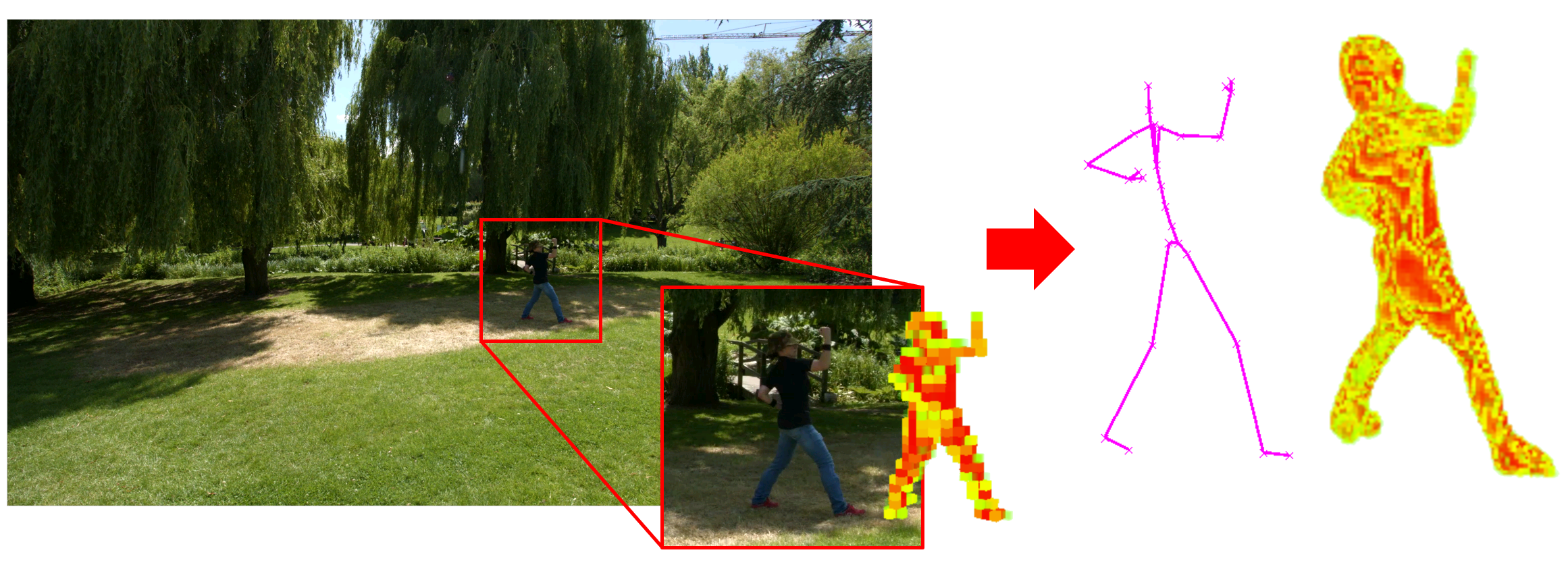}
\caption{Simultaneous estimation of 3D human pose and $4 \times$ upscaled volumetric body shape, from coarse visual hull data derived from a sparse set of wide-baseline views.}
\label{fig:teaser_totalcap}
\squeezeup
\squeezeup
\squeezeup
\end{figure}

\section{Introduction}

Multiple viewpoint video of open spaces (\eg for sports or surveillance) is often captured using a sparse set of wide-baseline static cameras, in which human subjects are relatively small (tens of pixels in height) due to their physical distance.  Nevertheless, it is useful to infer human behavioural data from this limited knowledge for performance analytics or security.  In this paper, we explore the possibility of using a deeply learned prior inferring high fidelity three-dimensional (3D) body shape and skeletal pose data from a coarse (low-resolution) volumetric estimate of body shape estimated across a sparse set of camera views (Fig.~\ref{fig:teaser_totalcap}).

The technical contribution of this paper is to explore the possibility of learning a deep representation for volumetric (3D) human body shape driven by a latent encoding for skeletal pose that can, in turn, be inferred from coarse volumetric shape data.  Specifically, we investigate whether convolutional autoencoder architectures, commonly applied to 2D visual content for de-noising and up-scaling (super-resolution),  may be adapted to up-scale volumetric 3D human shape whilst simultaneously providing high-level information on the 3D human pose from the bottle-neck (latent) representation of the autoencoder.   We propose a symmetric autoencoder with 3D convolutional stages capable of refining a probabilistic visual hull (PVH) \cite{Grauman2003} \ie voxel occupancy data derived at very coarse scale (grid resolution $32 \times 32 \times 32$ encompassing the subject).  We demonstrate that our autoencoder is able to estimate an up-scaled body shape volume at up to $128 \times 128 \times 128$ resolution, whilst able to estimate the skeleton joint positions of the subject to equal or better accuracy than the current state of the art methods due to deep learning.

\section{Related Work}

Our work makes a dual contribution to two long-standing Computer Vision problems: super-resolution (SR) and human pose estimation (HPE).

{\bf Super-resolution:} Data-driven approaches to image SR integrate pixel data \eg from auxiliary images \cite{Fattal2007}, or from a single image \cite{Glasner2009,Zhu2014}) to perform image up-scaling or restoration.  Model based approaches learn appearance priors from training images, applying these as optimization constraints to solve for SR content \cite{Freeman2002}. A wide variety of machine learning approaches have been applied to the latter \eg sparse coding \cite{Atalay2017}, regression trees~\cite{Schmidt2016}, and stacked autoencoders \cite{Vincent2008}; many such approaches are surveyed in \cite{srsurvey}.  Deep learning has more recently applied convolutional autoencoders for up-scaling of images \cite{Xie2012,Wang2015,Dong2016} and video \cite{Shi2016}; our work follows suit, extending symmetric autoencoders commonly used for image restoration to volumetric data using 3D (up-)convolutional layers \cite{Jain2008}.  Our work is not the first to propose volumetric super-resolution. Data-driven volumetric SR has been explored using multiple image fusion across the depth of field in \cite{Abrahamsson2017} and across multiple spectral channels in \cite{Atalay2017}.  Very recent work by Brock \etal explores deep variational auto-encoders for volumetric SR of objects~\cite{brock_generative_2016}. However, our work is unique in its ability to upscale to $4\times$ whilst simultaneously estimating human pose to a high accuracy, exploiting a learned latent representation encoding joint positions.

\begin{figure}[t!]
\centering
\includegraphics[width=1.0\linewidth]{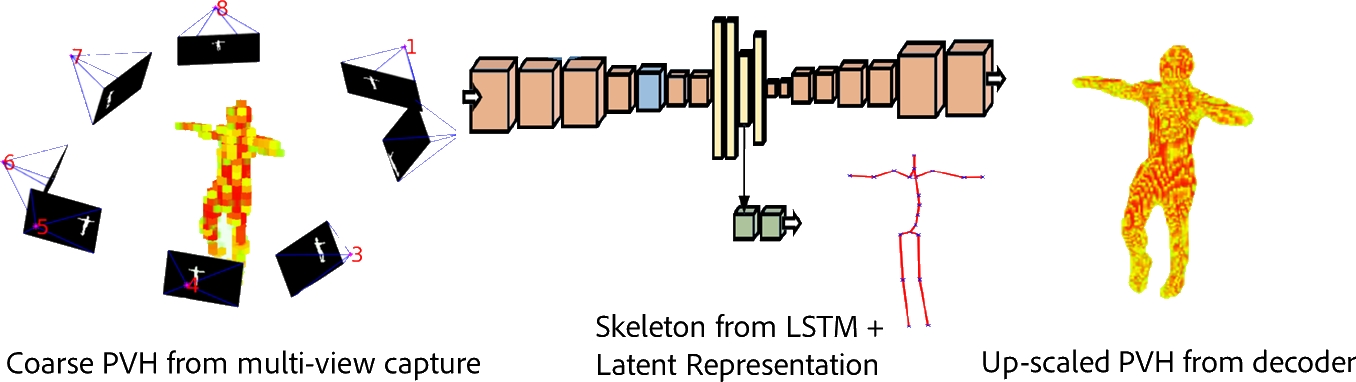}
\caption{Overview of the proposed method.  A coarse PVH is estimated as input volumetric data ($32^3$ voxels) and up-scaled via tricubic interpolation to a $(32n)^3$ voxel grid (where $n=\{1,2,4\}$). The input PVH is deeply encoded to the latent feature representation (3D joint positions). Non-linear decoding of the feature via successive up-convolutional layers yields a higher fidelity PVH of $(32n)^3$ voxels. }
\label{fig:overview}
\squeezeup
\end{figure}

{\bf Human pose estimation} has been classically approached through top-down fitting of models such as Pictorial structures~\cite{felzen03}, fused with Ada-Boost shape classification in \cite{andriluka09}.  Conditional dependencies between parts (limbs) during model fitting were explored in \cite{lan05,jiang09}.  Huang ~\cite{Huang2015} tracked 3D mesh deformation over time and attach a skeleton to tracked vertices.  The \emph{SMPL} body model~\cite{loper2015SMPL} provides a rich statistical body model that can be fitted to (possibly incomplete) visual data.  Marcard~\cite{SIP2017EG} explored the orthogonal modality of IMU measurements using SMPL for HPE without visual data. Malleson ~\cite{Malleson3DV17} used IMUs with a full kinematic solve to estimate 3D pose.  SMPL was recently applied to a deep encoder-decoder network to estimate 3D pose from 2D images~\cite{TanSMPLY2d3DBMVC17}.  Several deep approaches estimate 2D pose or infer 3D pose from intermediate 2D estimations.  DeepPose~\cite{Toshev2014} applies a convolutional neural network (CNN) cascade.  Descriptors learned via CNN have been used in 2D pose estimation from low-resolution 2D images \cite{Park2015} and real-time multi-subject 2D pose estimates were demonstrated by cao~\cite{cao2016realtimeCPM}. Sanzari~\cite{sanzari2016bayesianH36m} estimates the location of 2D joints, before predicting 3D pose using appearance and probable 3D pose of parts. Zhou~\cite{zhou2016sparsenessH36m} integrates 2D, 3D and temporal information to account for uncertainties in the data. 

The challenge of estimating 3D human pose from volumetric data is more sparsely explored. Trumble~\cite{TrumbleCVMP2DConvNet} used a spherical histogram and later voxel input to regress a pose estimate using a CNN~\cite{trumble_total_2017}. Pavlakos~\cite{pavlakos2017volumetricCVPR} used a simple volumetric representation in a 3D convnet for pose estimation. While Tekin~\cite{tekin2016structured} included a pretrained autoencoder within the network to enforce structural constraints.  Our work also trains an autoencoder for HPE but simultaneously infers a high resolution body model via a dual loss function.

\squeezeup
\section{Estimating Human Pose and Body Shape}
\squeezeup

Our method accepts a coarse resolution volumetric reconstruction of a subject as input, and in a single inference step estimates both the skeletal joint positions and a higher resolution (up-scaled) volumetric reconstruction of that subject (Fig.~\ref{fig:overview}).  Sec.~\ref{sec:pvh} first describes how the input volumetric reconstruction is formed, through a simplified form of Graumann 's probabilistic visual hull (PVH) \cite{Grauman2003}.  The architecture of our 3D convolutional autoencoder is then described in Sec.~\ref{sec:autoenc} including the dual loss function necessary to learn a deep representation of body shape and the latent pose representation.  Finally, Sec.~\ref{sec:train} describes the data augmentation and methodology for training the network.

\subsection {Volumetric Representation}
\label{sec:pvh}
The capture volume $\mathcal{V} \in \mathbb{R}^3$ containing the subject is observed by a set of $C$ calibrated cameras $c=\left[ 1,C \right]$ for which camera world position $T_c$ and orientation $R_c$ (both matrices in homogeneous form) are known as are intrinsics: camera focal length ($f_c$) and optical center $[o^x_c,o^y_c]$. An external process (\eg a person tracker) is assumed to isolate the bounding sub-volume $X_I \in \mathcal{V}$  corresponding to, and centered upon, a single subject of interest, and which is decimated to a coarse voxel grid $V=\{v_x^i, v_y^i, v_z^i\}$ for $i=[1,...,32^3]$ where $V$ denotes the coarse voxel volume passed as input to the network in Sec~\ref{sec:autoenc}. Each voxel $v^i \in V$ projects to coordinates $(x[v^i],y[v^i])$ in each camera view $c$ derived in homogeneous form via pin-hole projection:
\begin{eqnarray}
\left[\begin{array}{c}
\alpha x[v^i]\\
\alpha y[v^i]\\
\alpha
\end{array}\right]=
\left[\begin{array}{cccc}
f_c & 0 & o^x_c & 0 \\
0 & f_c & o^y_c & 0\\
0 & 0 & 1 &0 \\
\end{array}\right]
\left(-R_c^{-1}T_c\right)
\left[\begin{array}{c}
v_x^i\\
v_y^i\\
v_z^i\\ 
1
\end{array}\right].
\end{eqnarray}

Given a soft matte $I_c$ obtained, for example by background (clean-plate) subtraction, the probability of the voxel being part of the performer in a given view $c$ is: 
\begin{eqnarray}
p(v^i | c) = I_c(x[v^i],y[v^i]).  \label{eq:pvh1}
\end{eqnarray}
The overall probability of occupancy for a given voxel $p(v^i)$ is:
\begin{eqnarray}
p(v^i) = \prod_{i=1}^C 1/(1+e^{p(v^i|c)}). \label{eq:pvh4}
\end{eqnarray}
For all voxels $v^i \in V$ we compute $p(v^i)$ to form the coarse input PVH.

\subsection {Dual Loss Convolutional Autoencoder}
\label{sec:autoenc}

We use a convolutional autoencoder with a symmetrical `hourglass' (encoder-decoder) architecture.  The goal of the network is learn a deep representation given an input tensor $\mathbf{V_I} \in \mathbb{R}^{N \times N \times N \times 1}$ encoding the coarse PVH, $V$ at a given resolution $N=(32n)^3$, where $n=\{1,2,4\}$ is a configuration parameter determining the degree of up-scaling required from the network $(1\times,2\times,4\times)$ respectively.  The coarse PVH input $V$ is scaled via tri-cubic interpolation to fit $\mathbf{V_I}$.
\begin{figure}[t!]
\centering
\includegraphics[width=1.0\linewidth]{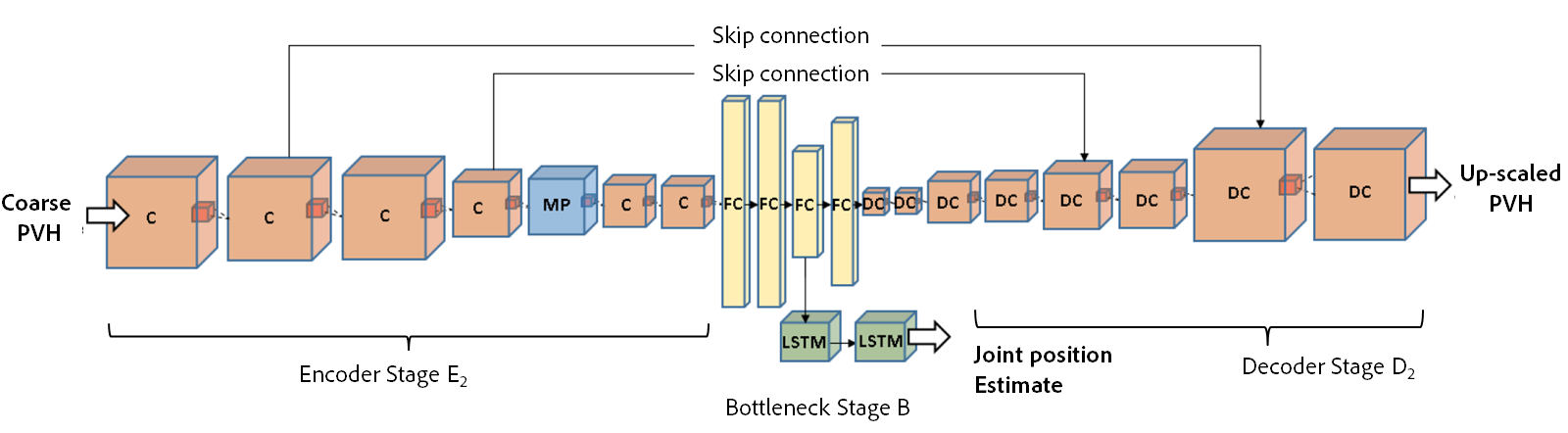}
\caption{Proposed convolutional autoencoder structure. The coarse input PVH is encoded into a latent feature representation via 3D (C)onvolutional, (M)ax-(P)ooling and (F)ully-(C)onnected layers.  The decoder uses the latent representation to synthesize an up-scaled PVH via (D)e-(C)onvolutional layers.  Two skip connections bridge the latent representation which is constrained during training to encode Cartesian joint positions. During inference these are passed through an LSTM to enhance temporal consistency to produce the joint position skeleton estimate.  Architecture pictured here is for $2\times$ scale-up -- in order to accommodate different receptive field sizes for $V_I/V_O$ (de-)convolutional layer count is adjusted -- see Tbl.~\ref{tbl:netparams}.
}
\label{fig:netarch}
\squeezeup
\end{figure}
We train the deep representation to solve the prediction problem $\mathbf{V_H} = \mathcal{F}(\mathbf{V_I})$ for similarly encoded output tensor $\mathbf{V_O}$, where 
\begin{eqnarray}
\mathbf{V_O} = \mathcal{F}(\mathbf{V_I})=\mathcal{D}(\mathcal{E}(\mathbf{V_I})) 
\end{eqnarray}
for the end to end trained encoder ($\mathcal{E}$) and decoder ($\mathcal{D}$) functions  The encoder yields a latent feature representation via a series of 3D convolutions, max-pooling and fully-connected layers.  We enforce $J(V_I)=\mathcal{E}(\mathbf{V_I})$ where $J(\mathbf{V_I})$ is a skeletal pose vector corresponding the input PVH; specifically a 78-D vector concatenation of $26\times$ 3D Cartesian joint coordinates in $\{x,y,z\}$.  The decoder half of the network inverts this process to output tensor $\mathbf{V_O}$ matching the input resolution but with higher fidelity content.
Fig.~\ref{fig:netarch} illustrates our architecture which incorporates two skip connections bypassing the network bottleneck to allow the output from a convolutional layer in the encoder to feed into the corresponding up-convolution layer in the decoder. Activations from the preceding layer in the main network and skip connection data are combined via mean average rather than channel augmentation/residuals. 

Tbl.~\ref{tbl:netparams} describes the parameters (filter count and size) of each layer.  We report experiments up-scaling to $n=\{1,2,4\}$ requiring varying sizes of receptive field to accommodate $\mathbf{V_I}$ and $\mathbf{V_O}$.  For each step up in scale, we add a single additional convolutional layer to the encoder, and two additional de-convolutional layers to the decoder.  Max-pooling occurs always at the fourth convolutional layer, and the filter size is $3 \times 3 \times 3$ except for the first two and last two layers, where the filter size is $5 \times 5 \times 5$ .

Learning the end-to-end mapping from coarse PVH to both an up-scaled PVH and accurate 3D joint positions requires estimation of the weights $\phi$ in $\mathcal{F}$ represented by the convolutional and deconvolutional kernels.

Specifically, given a collection of $M$ training triplets $\{\hat{\mathbf{V_I}}, \hat{\mathbf{V_O}}, \hat{J}\}$, where $p^i \in \hat{\mathbf{V_I}}$ is voxel data from a coarse (input) PVH, $q^i \in \hat{\mathbf{V_O}}$ is voxel data of an ideal up-scaled PVH, and $j$ is a vector of ideal joint positions for the given volume. We minimize the Mean Squared Error (MSE) at the outputs of the bottleneck and decoder stages across $M=N \times N \times N$ voxels:
\begin{eqnarray}
\mathcal{L(\phi)} = \frac{1}{M}\sum^M_{i=1} \| \mathcal{F}(p^i: \phi) -q^i \|^2_2+\lambda \| \mathcal{E}(\hat{\mathbf{V_I}}: \phi) -j \|^2_2. \label{eq:DualLoss}
\end{eqnarray}

These training triplets are formed by extracting voxel volumes from exemplar multi-view video footage at resolution $N \times N \times N$ (yielding $\hat{\mathbf{V_O}}$ and the artificially down-sampling to  $32 \times 32 \times 32$ to yield $V$ (from which $\mathbf{V_I}$ is up-sampled via tri-cubic interpolation).  Human pose (joint positions) corresponding to the multi-view video frame is acquired using a commercial (Vicon Blade) human performance capture system run in parallel with video acquisition (such annotations are provided with the {\em TotalCapture} and {\em Human3.6M} datasets).

\begin{table}[t!]
\centering
{
\small
\begin{tabular}{ccl}
\hline
Network Stage   &  \#Layers & \#Channels/Layer \\
\hline
$\mathrm{E}_1$ & 5 & 96*~ 96*~  96~ 96-M~ 96\\
$\mathrm{E}_2$ & 6 & 32*~ 64*~ 96~ 96-M~ 96~ 96\\
$\mathrm{E}_4$ & 7 & 32*~ 32*~32~ 64-M~ 96~ 96~ 96 \\
\hline
$\mathrm{B}$ & 4 & 1024~ 1024~ 78-J ~ 216 \\
\hline
$\mathrm{D}_1$ & 6 & 96~ 96~ 96~ 96~ 64* ~ 1*\\
$\mathrm{D}_2$ & 8 & 96~ 96~ 96~ 96~ 64~ 64~32*~ 1*\\
$\mathrm{D}_4$ & 10 & 96~ 96~ 96~ 96~ 64~ 64~32~ 32~ 32*~ 1*\\
\hline
\end{tabular}
}
\caption{Convolution layer parameters for the encoder ($\mathrm{E}_n$), bottleneck ($\mathrm{B}$), and decoder ($\mathrm{D}_n$) stages for $n=\{1,2,4)\times$.  Suffix $-M$ indicates max-pooling. All $E_n$ and $D_n$ layers learn $3 \times 3 \times 3$ filters, except where indicated by $*$ filters are $5 \times 5 \times 5$. All $B$ layers are fully-connected including the latent representation (3D joint positions) suffixed $-J$.}
\label{tbl:netparams}
\squeezeup
\squeezeup
\end{table}

\subsection {Training Methodology}
\label{sec:train}

To train $\mathcal{F}$ we use Adadelta ~\cite{zeiler2012adadelta} an extension of Adagrad, with the pose term of the dual loss (eq.~\ref{eq:DualLoss}) scaled by a factor of $\lambda$. We found the approach insensitive to this parameter up to an order of magnitude setting $\lambda=10^{-3}$ for all experiments. Below $10^{-3}$, the bottleneck convergences to a semantic representation of the pose that is stable but does not resemble joint angles --- above $10^{-2}$ the network will not converge. Data is augmented during training with a random rotation around the central vertical axis of the PVH. Before full network training, the encoder stage is trained separately, purely as a pose regression task, up to the latent representation layer. These trained weights initialize the encoder stage to help constrain the latent representation during full, dual-loss network training. Training typically converges within 100 epochs.
\begin{figure}[htb]
\centering
\includegraphics[width=1\linewidth]{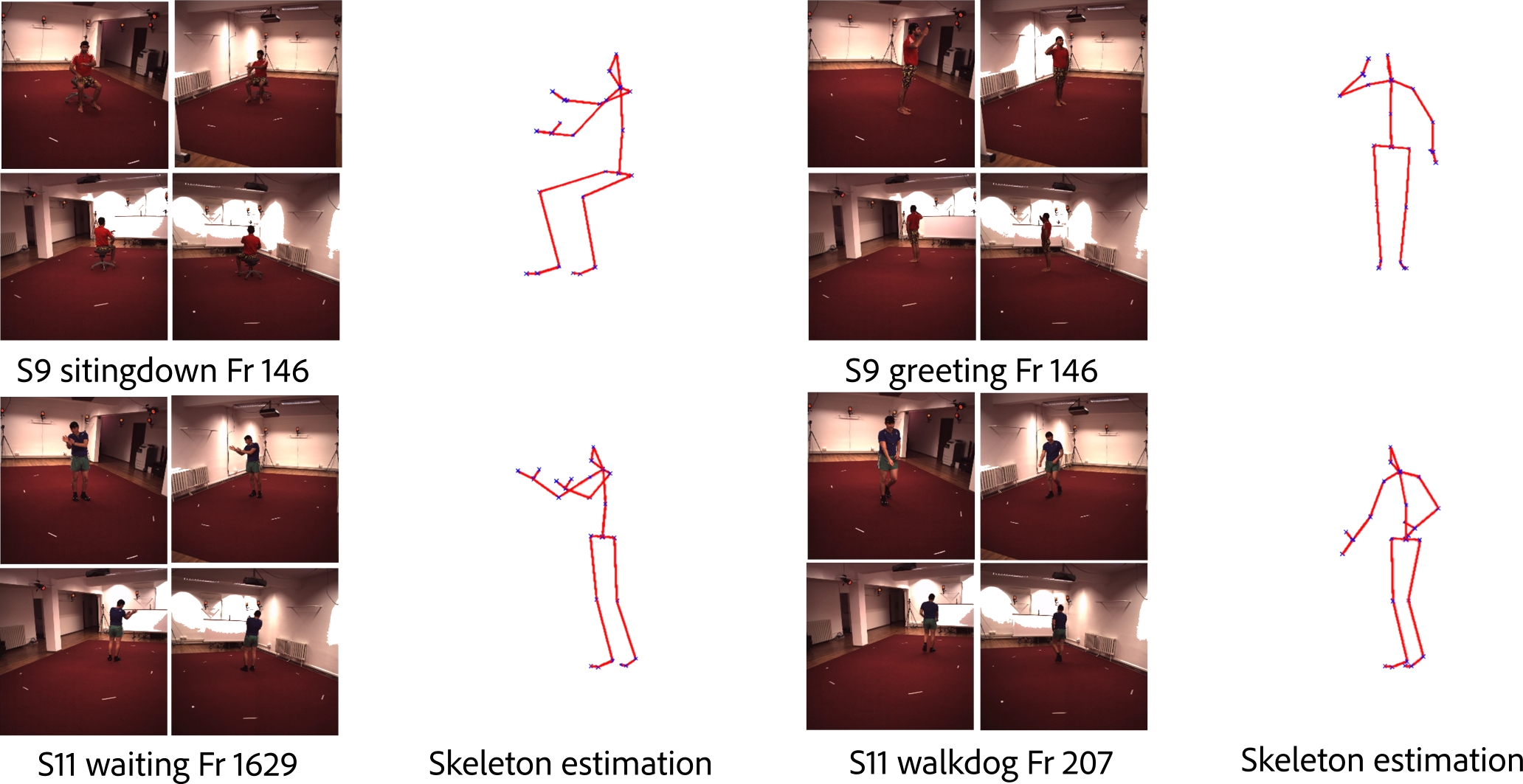}
\caption{Representative visual results for pose estimation on Human 3.6M across four test sequences (source footage from four views and inferred 3D skeletal pose). }
\label{fig:Human36MQual}
\end{figure}
\subsection {Enforcing Temporal Consistency}
\label{sec:LSTM}
Given the rich temporal nature of the pose sequences, it is prudent to exploit and enforce the temporal consistency of the otherwise detection based human joint estimation. By enforcing temporal consistency it is possible to smooth noise in individual joint detections that otherwise would cause large estimation errors. To learn a model of the motion over time we employ Long Short Term Memory (LSTM) layers~\cite{hochreiter1997LSTM}, they have been heavily utilized in applications where long term temporal correlation can be exploited such as \emph{e.g.} speech recognition~\cite{sak2014lstmSpeech}, video description~\cite{donahue2015lstmVideo}, and pose estimation~\cite{luo2017lstm}. LSTM layers are based on a  recurrent neural network (RNN). They can store and access information over long periods of time but are able to mitigate the vanishing gradient problem common in RNNs through a specialized gating mechanism. The input vector from the encoder $\mathbf{J}_i(t)=\mathcal{E}(\mathbf{V_I})$ at time $t$ consisting of concatenated joint spatial coordinates is passed through a series of gates resulting in an output joint vector $\mathbf{J}_o(t)$. The aim is to learn the function that minimizes the loss between the input vector and the output vector $\mathbf{J}_o = o_t \circ tanh(c_t)$ ($\circ$  denotes the Hadamard product) where $o_t$ is the output gate, and $c_t$ is the memory cell, a combination of the previous memory $c_{t-1}$ multiplied by a decay based forget gate, and the input gate. Thus, intuitively the LSTM result is the combination of the previous memory and the new input vector. In the implementation, our model consists of two LSTM layers both with 1024 memory cells, using a look back of $f=5$. 

\section{Evaluation and Discussion}
\label{sec:Eval}
To quantify the improvement in both the upscaling of low resolution volumetric representations and human pose estimation, we evaluate over three public multi-view video datasets of human actions. For \emph{Human 3.6M}~\cite{h36m_pami} we estimate the 3D human pose, and examine the performance of the skeleton estimation and volume upscaling in the \emph{TotalCapture}~\cite{trumble_total_2017} dataset. Finally, we visualize the results of the skeleton estimation and upscaling on the dataset \emph{TotalCaptureOutdoor}~\cite{Malleson:3DV:2017}, a challenging collection of multi-view human actions shot outdoors.

\subsection{Human 3.6M evaluation}
\begin{table}[htb]

\centering
{
\small
\begin{tabular}{lcccccccc}
\hline
Approach                              & Direct. & Discus & Eat & Greet. &Phone &Photo &Pose &Purch. \\ \hline
Lin~\cite{li2015maximumH36m}          & 132.7& 183.6 & 132.4& 164.4 &162.1&205.9&150.6 &171.3 \\
ekin~\cite{tekin2016fusingH36m}       & 85.0 & 108.8 & 84.4 & 98.9 &119.4&95.7 &98.5 &93.8 \\
Tome~\cite{tome2017liftingH36m}       &65.0 & 73.5 & 76.8 & 86.4 & 86.3&110.7&68.9 &74.8 \\ 
Trumble~\cite{trumble_total_2017}     & 92.7& 85.9& 72.3& 93.2& 86.2&101.2 &75.1 &78.0 \\ 
Lin~\cite{lin2017CVPRRPSM}              &58.0    & 68.3 & 63.3 &65.8  & 75.3 &93.1&61.2&65.7 \\ 
Martinez~\cite{martinez_simple_2017}          &51.8 & 56.2&58.1 & 59.0& 69.5&78.4  & 55.2 & 58.1 \\\hline

Proposed                              & 41.7& 43.2& 52.9& 70.0& 64.9& 83.0& 57.3& 63.5\\ \hline

& Sit. & Sit D & Smke & Wait &W.Dog& walk & W. toget. &Mean\\\hline
Lin~\cite{li2015maximumH36m} & 151.6 & 243.0 & 162.1 &170.7 &177.1& 96.6 & 127.9 & 162.1 \\
ekin~\cite{tekin2016fusingH36m} & 73.8 & 170.4 &85.1 &116.9 &113.7& 62.1 & 94.8 & 100.1 \\
Tome~\cite{tome2017liftingH36m} & 110.2 &173.9 &85.0 &85.8 &86.3 &71.4 &73.1 & 88.4 \\ 
Trumble~\cite{trumble_total_2017}  & 83.5& 94.8& 85.8&82.0 &114.6 &94.9 &79.7 &87.3 \\
Lin~\cite{lin2017CVPRRPSM}              &98.7 &127.7 &70.4 &68.2   & 73.0 & 50.6 & 57.7 & 73.1 \\ 
Martinez~\cite{martinez_simple_2017}      & 74.0 &94.6&62.3&59.1&65.1   &49.5 &52.4 &62.9 \\\hline
Proposed &61.0 &95.0 &70.0 &62.3 &66.2 &53.7 &52.4 &62.5 \\     \hline

\end{tabular}}
\caption{A Comparison of our approach to other works on the Human 3.6m dataset}
\squeezeup
\squeezeup
\label{tab:H36mResults}

\end{table}
The 3D human pose estimation dataset {\em Human3.6M}~\cite{h36m_pami} is a 4 camera view dataset of 10 subjects performing 210 actions at 50Hz in a $360^\circ$ arrangement. A 3D ground truth for joint positions (key points) are available via annotation using  a commercial marker-based motion capture system, allowing quantification of error. The dataset consists of 3.6 million video frames, balanced over 5 female and 6 male subjects. They perform common activities such as posing, sitting and giving directions. To allow comparison to other approaches we follow the same data partition protocol as in previous works ~\cite{h36m_pami,li2015maximumH36m,tekin2016fusingH36m,lin2017CVPRRPSM,tome2017liftingH36m,martinez_simple_2017}, and we use the publicly released foreground mattes. 
The training data consists of subjects S1, S5, S6, S7, S8 and it is tested on unseen subjects S9, S11.  We compare our approach to many previously published state of the art methods, using 3D Euclidean ($\mathrm{L}_2$) error  to compute accuracy. Error is measured between each ground truth and estimated 3D joint position and is averaged over all frames and all 17 joints in millimeters (mm). The results of our approach  are evaluated  qualitatively in Fig~\ref{fig:Human36MQual} and quantitatively in Tbl.~\ref{tab:H36mResults}, drawing comparison to state of the art approaches.

Our approach outperforms with the lowest mean joint error on the challenging Human3.6M dataset, slightly reduced over the state of the art approach by Martinez~\cite{martinez_simple_2017}, with a similar mean joint error of just over 6cm. This is averaged over both test subjects and the 59 sequences. The error decrease over the other approaches is possible due to the dual loss formulation ensuring that the skeleton is kept bounded by realistic 3D volume representations after the decoder. Our approach struggles with the actions Sit Down and Photo, the action sit down contains a chair and given the already poor quality of the PVH it is likely that such incorrect joint estimations occur. In the sequences of \emph{photo} the hands of the subject are close the subject head and it is likely the PVH volume doesn't contain enough discriminative information to correctly estimate their location. However, despite these two sequences, all others have a low error score and are smooth and qualitatively realistic. We show qualitative comparisons with respect to the ground truth in Fig.~\ref{fig:Human36MQual}. To illustrate the stability of our approach across different test subjects we performed five rounds of cross-validation using multiple pairs of test subjects with the remaining subjects held out for training the model.
\begin{table}[htb]

\centering
{
\small
\begin{tabular}{lcccccccc}
\hline
Approach      & Direct. & Discus & Eat & Greet. &Phone &Photo &Pose &Purch. \\ \hline

CrossVal mean & 52.2    & 49.8   & 53.0&63.1    &61.4  &76.8  &63.2 &59.3   \\
CrossVal sd   & 7.6     & 5.1    & 9.1 &5.8     &3.9   &4.7   &10.4 & 6.9   \\ \hline
Proposed      & 41.7& 43.2& 52.9& 70.0& 64.9& 83.0& 57.3& 63.5\\ \hline

              & Sit. & Sit D & Smke & Wait &W.Dog& walk & W. toget. &Mean\\\hline
CrossVal mean & 64.9 & 108.3 &68.9  &63.0  &63.6 &57.4  &55.0       &70.2  \\
CrossVal sd   & 5.2  &15.8   &5.7   &3.2   &6.9  &5.2   &3.0        &3.3\\ \hline
Proposed      &61.0 &95.0 &70.0 &62.3 &66.2 &53.7 &52.4 &62.5 \\     \hline

\end{tabular}}
\caption{A Comparison of testing on subjects S9 and S11 against a five-fold cross validation of other subject pairs on the Human 3.6m dataset}
\label{tab:H36mResultsCrossVal}
\squeezeup
\squeezeup
\end{table}
Table~\ref{tab:H36mResultsCrossVal} shows the standard test on S9 and S11 (mean accuracy of 62.5mm) from Table~\ref{tab:H36mResults} against the mean and standard deviation from our cross-validation experiment. The mean performance across random pairs of test subjects is similar to that of the official S9/S11 test split, and the $\sigma$ is low.  Thus they serve to show stability of the approach across different test subject pairings. 

\subsection{TotalCapture evaluation}
\begin{figure}[htb]
\centering
\includegraphics[width=1\linewidth]{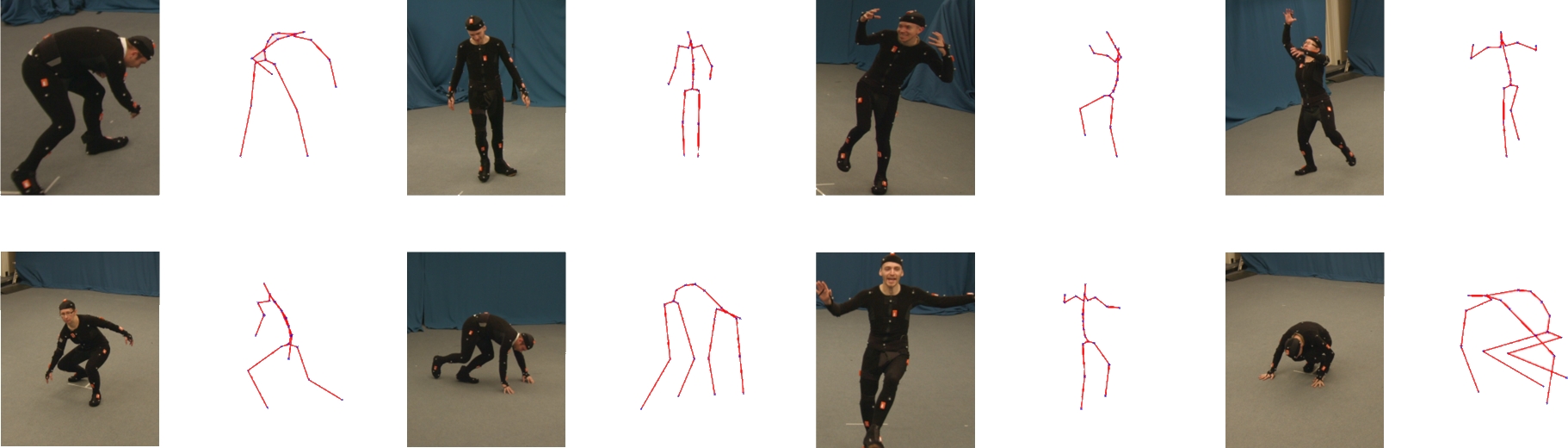}
\caption{Representative visual results from {\em TotalCapture} showing 3D pose estimation ($\times 2$ up-scaling). See Tbl. \ref{tab:totalcaptureResults} for quantitative results.}
\label{fig:TCexampleSkels}
\end{figure}

\begin{table}[htb]
\centering
{
\small
\begin{tabular}{lccccccc}
\hline
Approach &\multicolumn{3}{c}{SeenSubjects(S1,2,3)}&\multicolumn{3}{c}{UnseenSubjects(S4,5)} & Mean \\
                                        & W2 & FS3 & A3 & W2 & FS3 & A3 & \\ \hline
Tri-CPM-LSTM~\cite{cao2016realtimeCPM}       & 45.7 &102.8 & 71.9& 57.8 & 142.9 & 59.6 & 80.1 \\ 
2D Matte-LSTM~\cite{TrumbleCVMP2DConvNet}& 94.1 &128.9  &105.3 & 109.1& 168.5&120.6&121.1 \\ 
Trumble~\cite{trumble_total_2017} & 30.0 & 90.6 & 49.0 & 36.0 & 112.1 & 109.2 & 70.0 \\ \hline
AutoEnc-Front-Half                   & 42.0 & 120.5 & 59.8 & 58.4 & 162.1 & 103.4 & 85.4 \\
AutoEnc-x1-LSTM                    & 15.1 & 54.8 & 26.6 & 25.9 & 76.0 & 42.7 & 38.6 \\ 
AutoEnc-x2-LSTM                    & 13.0 & 47.0 & 23.0 & 21.8 & 68.5 & 40.9 & 34.1 \\ 
AutoEnc-x4-LSTM                    & 13.4 & 49.8 & 24.3 & 22.0 & 71.7 & 40.7 & 35.5 \\ \hline
\end{tabular}
}
\caption{Comparison of our approach on  TotalCapture  to other human pose estimation approaches, expressed as average per joint error (mm).}
\label{tab:totalcaptureResults}
\squeezeup
\squeezeup
\end{table}
In addition, we evaluate our approach on the TotalCapture dataset~\cite{trumble_total_2017}. This is also a 3D human pose estimation dataset with the ground truth joint position provided by Vicon markers. It is also captured indoors in a volume roughly measuring 8x4m with 8 calibration HD video cameras at 60Hz in a 360$^{\circ}$. There are a total of 5 subjects performing 4 actions with 3 repetitions at 60Hz in a $360^\circ$ arrangement. There are publicly released foreground mattes that we use as the input to our approach. Note to provide the Vicon groundtruth the subjects in both TotalCapture and Human3.6M are wearing dots visible to infrared cameras. However these dots are not used explicitly by our algorithm, and their size is negligible compared to the performance volume. There are five subjects in the dataset, four male, and one female, each performs four diverse performances, that are repeated 3 times: \emph{ROM, Walking, Acting, and Freestyle}. The length of each sequence is between 3000-5000 frames, this results in a total of $\sim 1.9M$ frames of synchronized groundtruth video data. Especially within the acting and freestyle sequences, there is great diversity in the actions performed, as illustrated in the qualitative results in Fig.~\ref{fig:TCexampleSkels}. 
To allow for comparison between seen and unseen subjects in the test evaluation, the test consists of sequences Freestyle3 (\textbf{FS3}), Acting (\textbf{A3}) and Walking2 (\textbf{W2}) on subjects 1,2,3,4 and 5. While the training is performed using the sequences of ROM1,2,3; Walking1,3; Freestyle1,2 and Acting1,2  on subjects 1, 2 and 3. We compare the pose estimation error for a number of upscale models; x1, x2, and x4 upscaling of the input PVH. Thus at the largest upscaling the PVH volume vector is $v \in  \mathbb{R}^{128 \times 128 \times 128}$. Tbl.~\ref{tab:totalcaptureResults} shows the results of the different upscaling models against the previous state of the art for the dataset.

All three learnt upscaling models reduce the mean error of the joints by over 50\% compared to previously published works for this dataset, with the error for some subjects sequences being reduced by an order of magnitude. Figure~\ref{fig:TCexampleSkels} provides some examples of the actions performed by the subjects and the excellent ability of the approach to estimate the pose.

\begin{figure}[htb]
\centering
\includegraphics[width=0.7\linewidth]{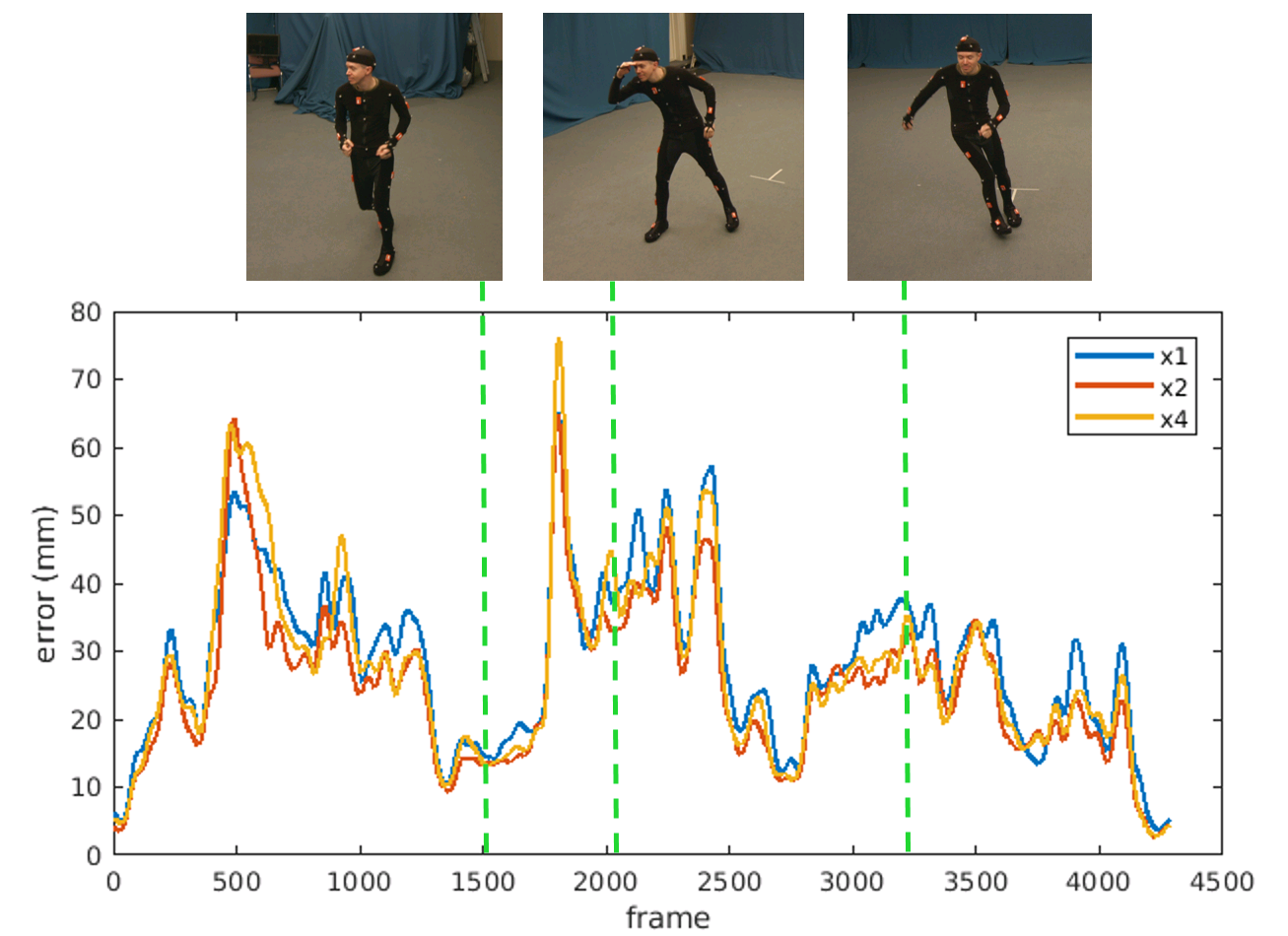}
\caption{Per frame skeletal error millimetres (mm) per joint on subject {\em S3 A3} in the {\em TotalCapture} test sequence.}
\label{fig:TotalcaptureFramebyFrame}
\end{figure}

\begin{figure}[htb]
\centering
\includegraphics[width=1\linewidth]{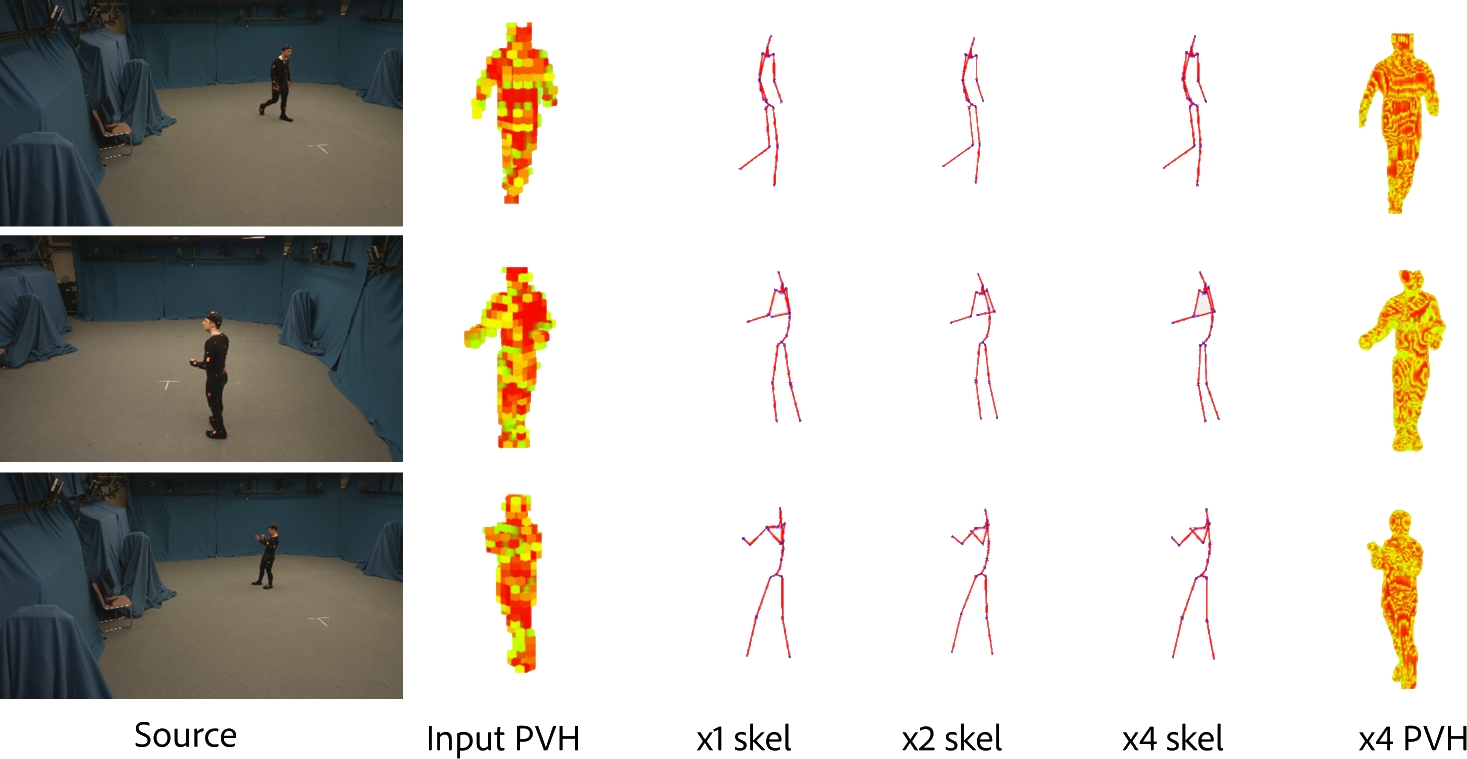}
\caption{Results illustrating the $\times 1, \times 2, \times 4$ upscaled volume for a representative coarse PVH alongside  upscaling inferred  skeletons.}
\label{fig:TCexampleAllSkels}
\squeezeup
\end{figure}

\begin{figure}[htb]
\centering
\includegraphics[width=0.8\linewidth]{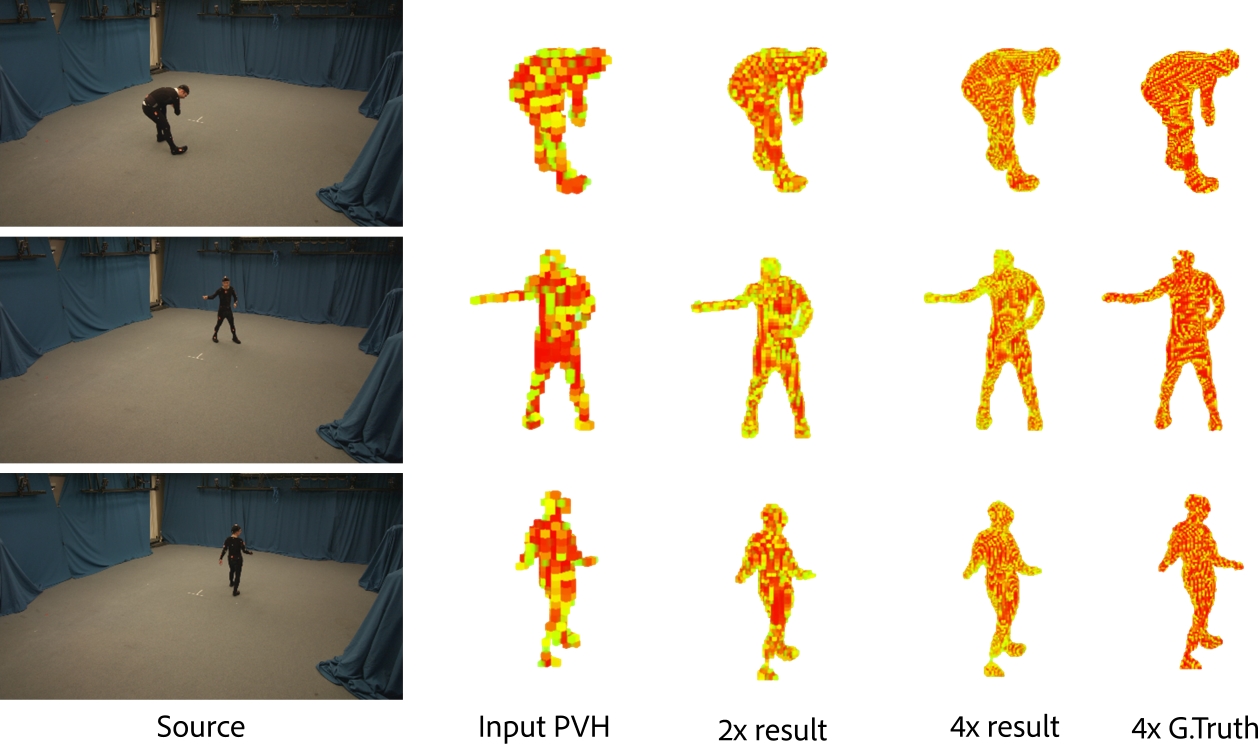}
\caption{Illustration of the upscaling ability of our approach on the TotalCapture dataset together with the native 128x128x128 groundtruth PVH}
\label{fig:UpscaleComp}
\squeezeup
\squeezeup
\end{figure}

Also, the table presents the \emph{AutoEnc-FrontHalf} results, this is shows initial convolutional encoder, without the decoder loss constraints. It provides a far higher error measure, indicating the importance of the dual loss constraining the skeleton pose space during training and inference. It is possible to examine the per frame error for subject 3, sequence Acting3, in Fig~\ref{fig:TotalcaptureFramebyFrame}. This figure shows how consistently low the error is across the full sequence. despite a number of challenging poses being performed by the actor.
There are a few error peaks, especially at the center point, and these are generally caused by a failure in the background segmentation from which the input PVH is generated, resulting in, for example, missing or weakly defined limb extremities.This data is under-represented within the training data. However otherwise error is low. Use of the symmetrical network and dual loss has provided a large reduction in joint error for the skeleton it is also possible to upscale the initially very coarse and small volume at up to $4 \times$ times. Figure~\ref{fig:TCexampleAllSkels} displays the initial volume estimate, the 4x upscaled volume and the skeleton estimate for 1x, 2x and 4x for a selection of example frames on the {\em TotalCapture} dataset.
The pose estimate for each upscaled model ($1 \times$,$2 \times$ and $4 \times$) is nearly identical as born out by the results previously presented in Tbl.~\ref{tab:totalcaptureResults}. however, the volume enhancement from the $4 \times$ upscaling is impressive allowing for greater details to be hallucinated without noise or degeneration. Tbl.~\ref{tab:totalcapturePVHResults} compares the input and output PVH volumes against a groundtruth high resolution volume generated directly from the camera views. The input volume is a naive tricubic upsampled volume and the error metric is MSE.
\begin{table}[htb]
\centering
{
\small
\begin{tabular}{lccccccc}
\hline
Approach                     &\multicolumn{3}{c}{SeenSubjects(S1,2,3)}&\multicolumn{3}{c}{UnseenSubjects(S4,5)} & Mean \\
                             & W2 & FS3 & A3 & W2 & FS3 & A3 & \\ \hline
AutoEnc-x2 input    &9.27 & 10.14 & 9.65 & 9.80 & 10.66 & 10.21 & 9.88\\ 
AutoEnc-x2 output    &0.34 & 0.37 & 0.34 & 0.40 & 0.46 & 0.39 & 0.37\\
AutoEnc-x4 input    &9.83 & 10.83 & 10.19 & 10.64 & 11.45 & 11.03 & 10.56\\ 
AutoEnc-x4 output    &0.50 & 0.55 & 0.50 & 0.58 & 0.68 & 0.59 & 0.56\\ \hline
\end{tabular}
}
\caption{Accuracy of generated volumes compared to tri-cubic upsampled input, over TotalCapture dataset. Expressed as mean voxel squared error $\times 10^{-3}$ from ground truth high resolution volume}
\label{tab:totalcapturePVHResults}
\squeezeup
\squeezeup
\end{table}
The table shows that an order of magnitude improvement occurs using our proposed method against a naive tricubic up-sampling method. Comparing the x2 and x4 outputs, the MSE increases only slightly despite the generative doubling of the actor volume. An illustration of the upscaling performance is shown in Figure~\ref{fig:UpscaleComp}, where the input and output volumes at up to x4 upscaling are shown for the TotalCapture dataset.

\begin{figure}[!b]
\centering
\includegraphics[width=0.75\linewidth]{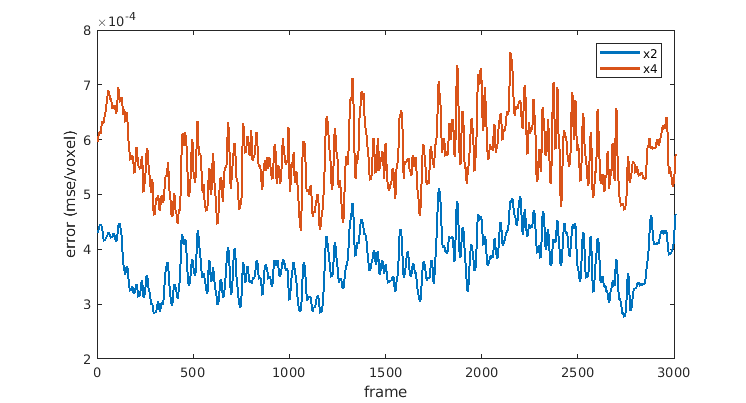}
\caption{Plotting volumetric reconstruction error per frame (MSE/voxel) on unseen subject {\em S4 A3} of the {\em TotalCapture} test sequence.}
\label{fig:VolFrame}
\squeezeup
\squeezeup
\end{figure}
Despite the initial block low-res PVH, we are able to accurately generate a hi-res PVHs at up to 4 times the size, that compare favorably to a natively generated (\ie $\mathbb{R}^{128 \times 128 \times 128}$) PVH. we are able to maintain extremity and no phantom volumes are formed in the upscaling process. Figure~\ref{fig:VolFrame} shows the per frame MSE over a sequence, for x2 and x4 upscaling. There is little difference between the two scales despite the greatly increased volume. Table ~\ref{tab:TotalCaptureTimings} shows the training and inference times (the latter near real-time) of our approach.
\begin{table}[b!]
\centering
{
\small
\begingroup
\setlength{\tabcolsep}{4pt} 
\begin{tabular}{cccccc}
\hline
&\multicolumn{2}{c}{Encoder Pre-train} &\multicolumn{2}{c}{Full Training}& Inference \\
PVH Scale &\makecell{Epochs to \\ converge} &minutes/epoch &\makecell{Epochs to \\ converge}  &minutes/epoch &millisec \\ \hline
x1        & 50 & 34 & 20 & 71 & 15 \\
x2        & 42 & 32 & 40 & 58 & 21 \\ 
x4        & 13 & 43 & 23 & 180 & 313 \\ \hline
\end{tabular}
\endgroup
\caption{Computational cost of model training and inference (TotalCapture dataset)}
\label{tab:TotalCaptureTimings}
}
\end{table}

\subsection{Outdoor footage evaluation}
\begin{figure}[htb]
\centering
\includegraphics[width=0.8\linewidth]{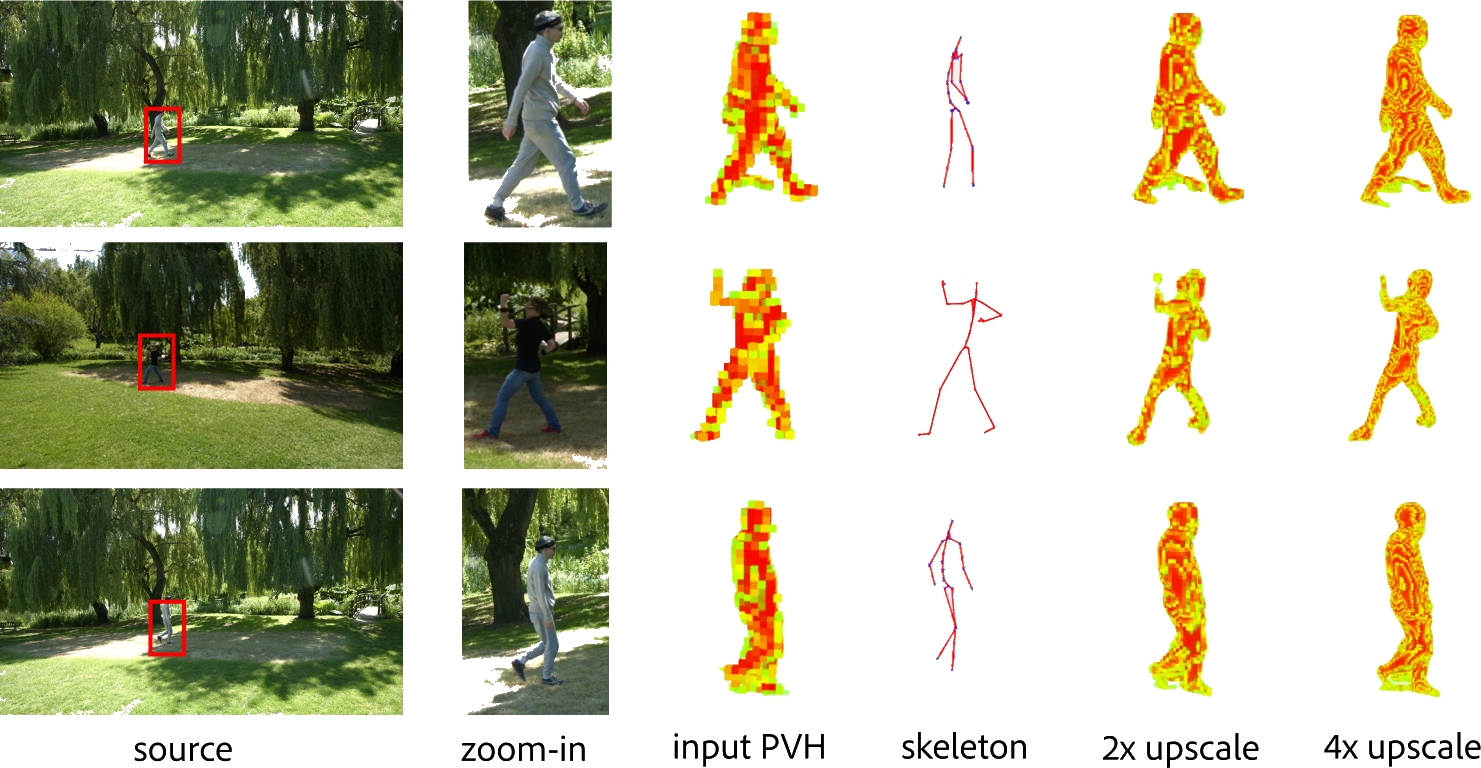}
\caption{Representative TotalCaptureOutdoor results showing  the low-res input PVH, and resulting skeleton and upscaled volumes}
\label{fig:OutdoorQual}
\squeezeup
\squeezeup
\end{figure}

\begin{figure}[htb]
\centering
\includegraphics[width=1\linewidth]{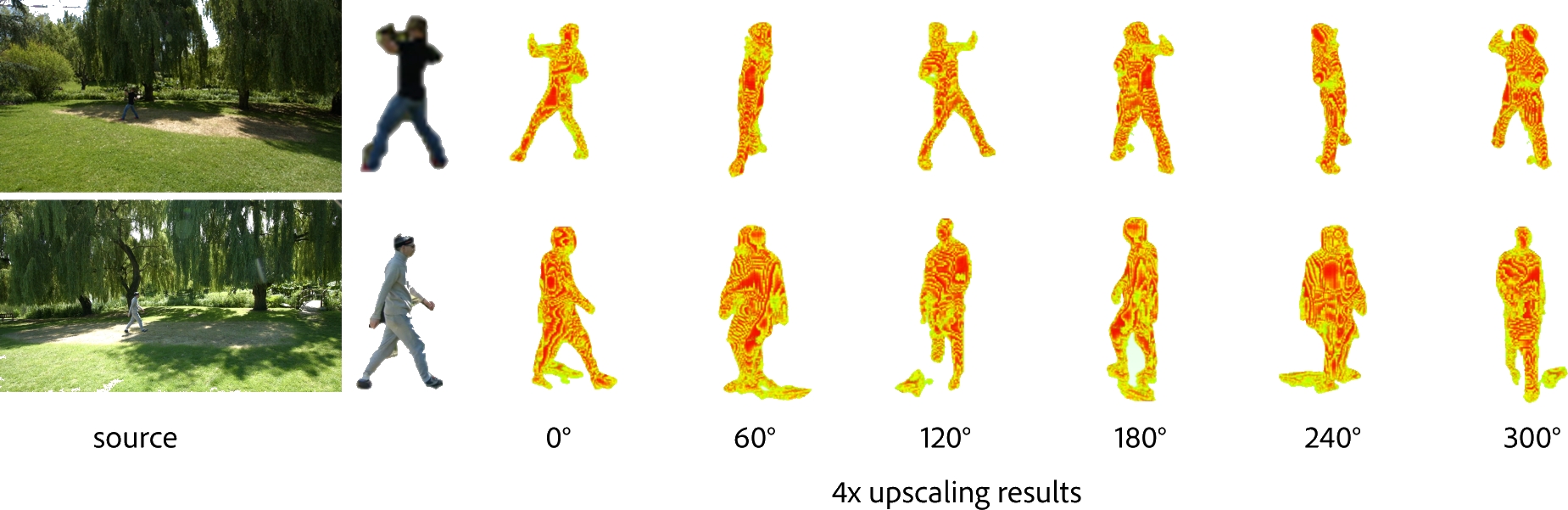}
\caption{Visualising the upscaled volumes from novel viewpoints.  3D reconstruction is of high quality despite the input PVH being captured from just two viewpoints.}
\label{fig:TCO360}
\end{figure}

To further demonstrate the flexibility of our upscaling and pose estimation approach, we test on a recent challenging dataset, \emph{TotalCaptureOutdoor}~\cite{Malleson3DV17}. This is an multi-view video dataset captured outdoors in a challenging uncontrolled conditions with a moving and varying background of trees and differing illumination. There are 6 video cameras placed in a 120$^\circ$ arrangement around the subject, with a large 10x10m capture volume used. This large capture volume means the subjects are small in the scene as shown in Figure~\ref{fig:TCoutdoorviews} below. For this dataset there are no released mattes, therefore we background subtraction was performed as a per pixel difference in HSV colour space to provide robust invariance against illumination change.
\begin{figure}[htb]
\centering
\includegraphics[width=1\linewidth]{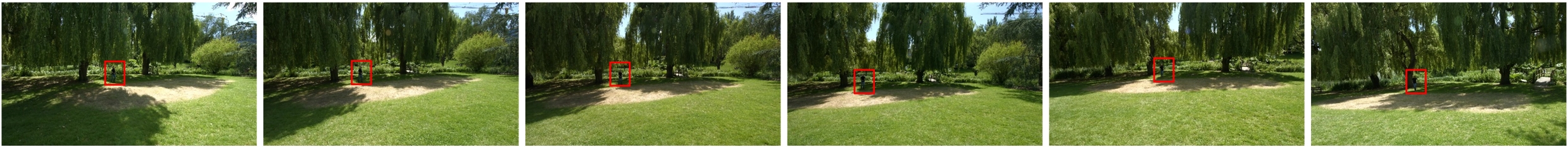}
\caption{{\em TotalCaptureOutdoor} dataset; red box indicates the person in the scene.}
\label{fig:TCoutdoorviews}
\end{figure}
There is no groundtruth annotation available for \emph{TotalCaptureOutdoor}, however, we are present several illustrative results on two sequences: {\em Subject1, Freestyle, and Acting1}. Given the small size of the subjects, a traditional 3D pose estimation or volume reconstruction would be challenging. However as shown in Figure~\ref{fig:OutdoorQual} we are able to use a small blocky low resolution PVH volume, that is upscaled by a factor of $ \times 4$ to produce a smooth approximation of the distant subject together with an accurate estimation of their joints. 
Furthermore, despite the camera being arranged in a 120$^\circ$ arc, we are able to simulate novel viewpoints of the upscaled full volume as shown in Figure~\ref{fig:TCO360}, where complete 360 views are possible. This upscaling enables a future avenue of work, creating a  3D model of the upscaled volume to produce VR/AR compositions or for film/ sports post-production.

\section{Conclusion}

We proposed a deep representation for volumetric (3D) human body shape driven by a latent encoding for the skeletal pose that can, in turn, be inferred from very coarse ($\mathbb{R}^{32 \times 32 \times 32}$) volumetric shape data.  In a single inference pass our convolutional autoencoder both up-scales up the provided volumetric data (demonstrated to a factor of $4 \times$) and predicts 3D human pose (joint positions) with greater or equal accuracy to state of the art deep human pose estimation approaches.  

Future work could explore the end-to-end integration of the LSTM to the autoencoder during training since the latter currently learns no temporal prior to aid pose or volume regression.  Nevertheless, we achieve state of the art results on very low resolution volumetric input, indicating the technique has potential to enable behavioural analytics using multi-view video footage shot at a distance.

\section*{Acknowledgements}
The work was supported by an EPSRC doctoral bursary and InnovateUK via the TotalCapture project, grant agreement 102685.   The work was supported in part through the donation of GPU hardware by the NVidia corporation.

\bibliographystyle{splncs}
\bibliography{pose}
\end{document}